\documentclass[12pt]{article}
\usepackage{graphicx}
\usepackage{amsmath}
\usepackage{bm}
\usepackage{authblk}
 \usepackage[utf8]{inputenc}

\begin{document}

\title{Random thoughts about  Complexity,  Data and  Models}

\author{Hykel Hosni $^{1}$
and Angelo Vulpiani $^{2}$ \vspace{2mm} \\
$^{1}$  Dipartimento di Filosofia, Universit\`a di Milano,
 Via Festa del Perdono 7, I- 20122 Milano, Italy. \texttt{hykel.hosni@unimi.it}
\\
$^{2}$ Dipartimento di Fisica, Universit\`a Sapienza, 
Piazzale Aldo Moro 2, I-00185 Roma, Italy. \texttt{angelo.vulpiani@roma1.infn.it}\vspace{2mm}}

\date{\today}

 \maketitle

 \begin{abstract}
   Data Science and Machine learning have been growing strong for the
   past decade. We argue that to make the most
   of this exciting field we should resist the temptation of assuming
   that forecasting can be reduced to brute-force data analytics. This
   owes to the fact that modelling, as we illustrate below, requires
   mastering the \emph{art} of selecting relevant variables.
 
   More specifically, we investigate   the subtle
   relation between ``data and models'' by focussing on the role
   played by algorithmic complexity, which contributed to making
   mathematically rigorous the long-standing idea that to understand
   empirical phenomena is to describe the rules which generate the
   data in terms which are ``simpler'' than the data itself.

   A key issue for the appraisal of the relation between algorithmic
   complexity and algorithmic learning is to do with a much needed
   clarification on the related but distinct concepts of
   compressibility, determinism and predictability. To this end we
   will illustrate that the evolution law of a chaotic system is
   compressibile, but a generic initial condition for it is not,
   making the time series generated by chaotic systems incompressible
   in general. Hence knowledge of the rules which govern an empirical
   phenomenon are not sufficient for predicting its outcomes. In turn
   this implies that there is more to understanding phenomena than
   learning -- even from data alone -- such rules. This can be
   achieved only in those cases when we are capable of ``good
   modelling''.

   Clearly, the very idea of algorithmic complexity rests on Turing's
   seminal analysis of computation. This motivates our remarks
   on this extremely telling example of analogy-based abstract
   modelling which is nonetheless heavily informed by empirical facts.
 
 \end{abstract}

\paragraph{KEYWORDS} Models, algorithmic complexity,  explanation, computation, big data.

\section{Introduction}
   \begin{flushright}
 \begin{minipage}[r]{0.8\linewidth}
     \begin{footnotesize}
       Suppose that we could find a finite system of rules which
       enabled us to say whether any given formula was demonstrable or
       not. This system would embody a theorem of
       metamathematics. There is of course no such theorem and this is
       very fortunate, since if there were we should have a mechanical
       set of rules for the solution of all mathematical problems, and
       our activities as mathematicians would come to an end. \cite{H29}
     \end{footnotesize}
 \end{minipage}
   \end{flushright}
 
Is data science proving what Hardy feared, but lightheartedly dismissed as
obviously false? For if
we take Hardy's ``finite system of rules'' to compress algorithmically
empirical data, and we take machine learning to be capable of using
this data to produce scientific knowledge, isn't there reason to
believe that Hardy was indeed wrong? 

It has been argued time and again in science and philosophy that the
aim of science  is to organise in the most economical fashion the data
collected from experiments. Mach \cite{M07,M14} championed
this view, which tuned out to be compelling to many scientists:
 
\begin{quote}
  The so-called descriptive sciences must chiefly remain content with
  reconstructing individual facts \ldots. But in sciences that are more
  highly developed, rules for the reconstruction of great numbers of
  facts may be embodied in a single expression.
\end{quote}

It is remarkable that recently such an approach has been reconsidered
in the framework of algorithmic complexity \cite{LV09}
from researchers without specific philosophical interests.  For
instance, Solomonoff, one of the fathers of the theory of the
algorithmic complexity, identifies
 a scientific law with an algorithm for compressing the results of 
 experiments \cite{S64}:
 
\begin{quote}
  The laws of science that have been discovered can be viewed as
  summaries of large amounts of empirical data about the universe. In
  the present context, each such law can be transformed into a method
  of compactly coding the empirical data that gave rise to the law.
\end{quote}

Similar opinions are shared by \cite{B07}, a scientist well know for his popular books:

\begin{quote}
  The intelligibility of the world amounts to the fact that we find it
  to be algorithmically compressible. We can replace sequences of
  facts and observational data by abbreviated statements which contain
  the same information content. These abbreviations we often call laws
  of Nature. If the world were not algorithmically compressible, then
  there would exist no simple laws of Nature.
\end{quote}

This view of the science, which puts at its center information and
algorithms, has unsurprisingly gained much interest and broad success in the last
decades. And this certainly created a favourable framework for the
recent rise of datacentric enthusiasm. We have put forward some
methodological words of caution about the temptation of replacing
the art of scientific modelling with the machine-aided analysis of time series \cite{CDH16},
more generally we have argued \cite{HV18} that
the accuracy of forecasts need not be monotonic in the amount of data,
-- indeed the opposite is true in paramount examples from dynamical
systems.

Our present concern is to do with a rather subtle but frequent
conceptual confusion which is likely to arise in many scientific contexts which refer to algorithmic complexity,
information and  chaos. Specifically it is a confusion on the relation among
the  determinism, predictability, and the view according to
which the scientific understanding of a
phenomenon can be acccounted for in terms of algorithmic compression. 

A telling example is provided for
instance,  by Davies \cite{D90}:
 
\begin{quote}
  there is a wide class of physical systems, the so-called chaotic
  ones, which are not algorithmically compressible.
\end{quote}

This sounds intuitively appealing, but it is misleading. As we will see in
Section 3 the evolution law of a chaotic system \emph{is}
compressible. What \emph{is not} compressible is the time sequence
generated by chaotic systems, and this, as argued in \cite{CV17},  is due to the non-compressibility 
of a generic initial condition.

Hence, the aim of the present paper is  to assess in some detail the
methodological scope and relevance for scientific enquiry of the
concepts of compression and algorithmic complexity. Before tackling our key question,  namely \emph{do scientific laws compress
  empirical data?}, it is worth recalling that the concepts of
compression and algorithmic complexity are direct offsprings of the
visionary intuitions which led  Alan Mathison Turing to provide an
abstract and very general model of computation, which in addition is
capable of identifying the bounds of and computability. Reviewing, if
very quickly, the
path followed by Turing will give us precious insights in the subtle
art of modelling and how this depends on the careful selection of
relevant data. 

\section{Alan Turing's model of computation}

The early 1930s have been dotted with deep results on the powers and
limitations of formal mathematical reasoning. Chief among them is Kurt
G\"odel's proof of the existence of undecidable sentences in first
order languages capable of expressing arithmetic. This marked a
distinction between what can be seen as mathematically true and
what is mathematically provable: if proofs are to be consistent, then
there exists true statements which can neither be proved nor
disproved. And if this was not enough, one such \emph{undecidable}
statement expresses the consistency of that very system of proof.

G\"odel's incompleteness theorems left logically open the possibility
that the \emph{Entscheidungsproblem} could be solved
positively for first order logic. As put by Hilbert and Ackermann 
\begin{quote}
  The Entscheidungsproblem is solved if one knows a procedure which
  will permit one to decide, using a finite number of operations, on
  the [satisfiability] of a given logical expression.
\end{quote}

Of course the incompleteness result cast serious doubts on a positive
solution, but to prove that the Entscheidungsproblem was indeed
unsolvable one first needed to state it in a mathematically rigorous
way rather than resting on an intuitive understanding of a
``procedure'' or ``algorithm''. This was the task that Turing gave to
\emph{computing machines} in  \cite{T36}  -- which 
have been called ever since \emph{Turing Machines} -- by means of which it became possible to suppose
that the informal concept of ``effective procedure'' (or 
``algorithm'') coincides with the class of functions
computable by means of a Turing Machine. Armed with this
formalisation, it was possible for Turing to prove that the
Entscheidungsproblem  was indeed unsolvable.  We are not, at present, so much
interested in either the proof or the consequences of this
result, which the interested reader can appreciate e.g. in
 \cite{D00}. Rather we are interested in recalling
Turing's initial \emph{motivation} and formulation for the definition
of algorithmic procedure in  \cite{T36}. In other words, how
Turing constructed his \emph{model} of computability. The reader who
wishes to find out more is referred to  the
unsurpassed analysis provided in  \cite{G95} by Turing's only student at Manchester.

Turing arrives at his model of computation in four steps:
\begin{enumerate}
\item Finds out what is the \emph{right question to ask};
\item Carries out an \emph{analogy} with the human activity of
  computing and \emph{observes} the cognitive limitations of humans as
  they carry out computations;
\item \emph{Claims} that the actions and operations which define a Turing
  Machine are sufficient to account for the concept of ``effective
  procedure'' i.e. for a mathematical definition of algorithms.
\end{enumerate}

Step 1) is of fundamental importance. The question as to whether all
solvable problems can be solved algorithmically is clearly too vast to
be tackled. But Turing's modelling genius allows him to see that there
was no loss of generality in asking \emph{which numbers are
  computable}, i.e ``real numbers whose expressions as a decimal are
calculable by finite means''. Then in step 2) Turing imagines a human
``computor'' tackling this task and strips away all the irrelevant
features from the picture. So to compute, one needs to read, write and
erase symbols on a large enough physical support (tape). And to do so
purposefully, the computor must perform those \emph{actions} according to the
instructions provided by their \emph{state of mind}. What cannot be stripped away are the
\emph{limitations} of the human computor: there is only a limited
number of symbols that can be read, erased and written at the same
time, and the state of mind of the computor may contain only a finite
number of instructions. Those abstractions and constraints give Turing
a model of computing behaviour which relied only on finite means and a
definite procedure. Then, step 3) is the claim that the class of
numbers (functions etc.) that can be
computed mechanically coincides with that computable by means of a
suitably defined Turing Machine. Within this model of computability
Turing proved the undecidability of the Halting Problem, which
establishes the existence of well defined problems which
cannot be solved algorithmically. 

A far reaching consequence of this is that not all data can be compressed
algorithmically. If scientific understanding were adequately modelled
by algorithmic compressibility, this would imply that there are unknowable scientific
facts. But what would that mean?

\section{Do scientific laws compress empirical data?}

Let us begin by noting the rather obvious fact that once a scientific
law has been established,  one has reached some sort compression. But
care must be placed in handling this conclusion. Especially when chaos
enters the picture. So let us follow Turing in delimiting the scope of
our question to two rather difficult and important problems which are useful to clarify  the role of chaos, initial conditions and 
algorithmic complexity in scientific practice: first 
chaotic deterministic systems and second the  features of fully developed turbulence (FDT) \cite{F95}.

\subsection{Classical mechanics and chaos}

As first we have to stress that 
there is a persistent confusion about determinism, chaos  and predictability. 
It is possible to understand that determinism  and predictability are completely unrelated. 
In few words we can say that determinism is ontic and has to do with how Nature behaves, 
while predictability is epistemic and is related to what the human beings are able to compute,
On the other hand a careful  analysis of chaotic systems in terms of
 the Lyapunov exponents and the Kolmogorov-Sinai entropy shows how deterministic chaos, 
although with an epistemic character, is non subjective at all, 
for a detailed discussion see (Atmanspacher 2002, Chibbaro et al 2014). 

It is well know that from the Newton equation
and the gravitation law one can derive many important astronomical facts,
for instance the Kepler's laws. From this, however it is not
completely  correct to conclude that  the Newton equation and the gravitation law are capable of compressing all astronomical
behavior. Indeed, after the seminal contribution of Poincar\'e, we
know that a system of three bodies  interacting with the gravitational
force,  usually is chaotic.

Rather than developing this difficult question in full detail, let us
consider a one dimensional map: 
$$
x_{t+1}=2 x_t \,\,\, mod \, 1 \, .
\eqno(1)
$$
This model, called \emph{Bernoulli shift}, is known to be chaotic, i.e. a small initial uncertainty increases exponentially, namely 
$$
\delta x_t \sim 2^t \delta x_0 \,.
\eqno(2)
$$

In spite of its (apparent) simplicity the Bernoulli shift 
shares with the three body problem many central features, so let us analyse the problem of the  compression,
with accuracy $\Delta$, of a sequence
$x_t, \, 0<t<T$, generated by the rule (1). In the parlance of
algorithmic information theory,  this is equivalent to  the
problem of transmitting the sequence to say, a friend.  At first glance, the problem seems quite simple: we could opt for
transmitting $x_0$ and the rule (1), which costs a number of bits
independent of $T$ . Our friend  would then be left with the task of generating the sequence 
$x_1, x_2, ... , x_T$.
However, we must also choose the number of bits to which $x_0$ should be specified.
From (2), the accuracy $\Delta$ at time $T$ requires accuracy 
$\delta_0 \sim 2^{-T} \Delta$  for $x_0$, hence
that the number of bits specifying $x_0$ grows with $T$. 
\\
Writing the initial condition in the forms
$$
x_0=\sum_{n=1}{a_n \over 2^n}=(a_1, a _2, .... )
\eqno(3)
$$
we understand that the evolution law of (1) is nothing but a shift of the binary point of
the sequence $\{ a_1, a _2, ..., \}$.
So  we have to tackle the
problem of the complexity of a sequence of symbols, 
$\{ a_0, a_1, ... \}$.

Recall that the evolution of $x_0$ is \emph{regular} (e.g. periodic) if its sequence
 $\{ a_1, a _2, ..., \}$ is not complex,  while it is \emph{irregular}
 if $\{ a_1, a_2, ..., \}$ cannot be compressed.

Therefore both in systems with regular behavior
(e.g. the pendulum) and in chaos (e.g. the three body or the
Bernoulli's shift) the evolution law is straightforward to
compress. The difference between the  two classes of  systems lies
with the outputs: which are always regular in the pendulum, whereas in
the Bernoulli's shift can be regular or irregular at varying initial conditions.

The  conclusion is that, in a deterministic system,  the
details of the time evolution are well hidden in the initial condition
which is, typically incompressibile.  This follows from an important result of Martin-L\"of 
who showed  that almost all infinite
binary sequences, which express the real numbers in $[0,1]$, are
complex \cite{M66}. Note that the identification randomness with
incompressibility put forward by Martin-L\"of (and further developed
by Kolmogorov and Chaitin) follows rather closely the modelling
footsteps of Turing's model of computation. Indeed it has become
standard to refer to the claim that ``incompressibility'' is the
mathematically rigorous counterpart of ``random'' as the
\emph{Martin-L\"of Thesis} \cite{LV09}.

\subsection{Turbulence}
Turbulent flows, a paradigmatic case of complex system, are governed 
by the Navier-Stokes equations (NSE) which can be written in two lines.
So, naively, one could conclude  that, since we know the equation for the time evolution
of the velocity field,  somehow, the phenomenon of turbulence has been compressed.
The study of   some specific  aspects of the turbulence allows for  the understanding  
of the precise meaning of such a conclusion.
As first we consider the   initial conditions,
of course in any experiment they  are necessarily known
with a limited precision.
A rather severe limitation is due to the fact that
in the limit of very large Reynolds numbers $R_e$
for a proper description of the turbulent velocity field  
it is necessary to consider a huge  number of variables
which increases quickly with $R_e$.
Therefore for the typical values of $R_e$ in the limit of fully developed turbulence (FDT), 
because of  the gigantic amount  of data necessary to describe  the involved degrees of freedom,
we have an obvious impossibility to access to the initial conditions with the  proper  accuracy.
\\
In addition at large $R_e$ the NSE are chaotic: the distance between two initially close initial conditions 
increase very fast.
Therefore as consequence of the practical impossibility to access  the initial conditions 
with high accuracy, and the presence of deterministic chaos,  even a very powerful computer and
 accurate numerical algorithms  is not possible  to perform a simulation of 
 the NSE  for a long term and compare the  prediction with the experimental results.
\\
Since the practical impossibility to compare the experimental results
with the numerical computation of the velocity field, 
 we cannot say that the NSE are able to compress the turbulent behaviors.
Nevertheless there is  a general consensus on the validity of the NSE  for the  FDT.
We can mention at least four items supporting the opinion that NSE are able to describe FDT,
the agreement of the results observed in FDT and those obtained by the  NSE  for:
\\
a) short time prediction of the velocity field;
\\
b) long time prediction of averaged (e.g. spatially caorse grained) quantities;
\\
c) the scaling laws, and more generally, the statistical features;
\\
d)] the qualitative and quantitative spatio-temporal features (e.g. large scale coherent structures).

\section{Science only from data?}

The NSE have been derived
on a theoretical basis  using the Newton equations and  assuming the
hypothesis of the continuity of matter plus some thermodynamic
considerations. In other words, by \emph{modelling}. One can wonder about the possibility to obtain
the  NSE just looking directly at experimental data.

A less ambitious task, but with the same conceptual aspects,
is   to  build models in finite dimension on the basis of experimental data \cite{GW94}.
 And again for sake of simplicity only, let us focus on the discrete
 time case.  The most favourable case is clearly the one in which we know that the state of the system  at time $k$ 
 is a finite dimensional vector ${\bf x}_k$. So let us start from here
 and consider the problem of making predictions from the available data,
i.e. a long time sequence. 

The natural approach is  to search for a past state similar to the present state of a given phenomenon of interest,
 then, looking at the sequence of events that followed the past state, one may infer
by analogy the evolution that will follow the present state \cite{CCFV12}.
In other words, given a
known sequence of \emph{analogues}, i.e. of past states ${\bf x}_1, ... , {\bf x}_M$  which resemble each other closely in pairs, so that
$|{\bf x}_k- {\bf x}_M|<\epsilon$    with $\epsilon$ reasonably small, one makes
the approximate prediction:
$$
{\bf x}_{M+1}={\bf x}_{k+1}
$$
if  ${\bf x}_k$  is an analogue of ${\bf x}_M$.

In the case the above protocol can be used,
one may then proceed to build a model of the phenomenon,
i.e. to determine a function ${\bf f}({\bf x})$ such that the sequences of states is well approximated
by the dynamical system 
$$
{\bf x}_{k+1}={\bf f}({\bf x}_k) \,\, . 
\eqno(4)
$$

 The application of this method requires
knowledge of at least one analogue. It is possible  to realise,
 -- and this is the Kac lemma --   that such knowledge
requires sufficiently long sequences, at least of duration of order 
 $ T_R \sim (L/\epsilon)^D$, where, if the system is dissipative,
 $D$ is the dimension of the attractor, basically  $D$ is the number of
 relevant variables.
 
The exponential growth of $T_R$ as a function of $D$
has a severe  impact  on our ability to make predictions,
and build the evolution law (4),
solely relying on previously acquired data. One can say that $D$ larger than 6 renders
the approach described here useless, because it makes it practically impossible to
observe the ''same'' state twice, i.e. within an acceptable accuracy $\epsilon$.
\\
On the other hand typically the  {\it state 
of the system}, i.e. the variables which describe the phenomenon 
under investigation, is not known.
Therefore an unavoidable technical aspect is the determination of proper state  of the system
from the study of a time series $\{ u_1, u_2, ..., u_M \}$, where $u$ is an observable.
The most relevant result for such a problem is due to Takens \cite{T81} 
who has been able to show that, at least from a mathematical point of view,
it is possible (if we know that the system is deterministic and is described by a finite dimensional vector) to determine
the proper variable ${\bf X}$.
In a nutshell: there is a finite integer $m$ such that the delay coordinate
vector (of dimension $m$)
$$
{\bf y}_k^{(m)}=(u_k, u_{k-1}, .. , u_{k-m+1})
$$
can faithfully reconstruct the properties of the underlying
dynamics, from  heuristic arguments one can expect $m= [D]+1$.
\\
Of course the practical limitation due to the exponential increasing of  $T_R$ as function of $D$,
is present also in the Takens's method; therefore we have   rather severe practical limitations.

A way out of this practical limitation, as we argued in  \cite{CCFV12,HV18}
consists a careful selection of the relevant features of the
phenomenon of interest, that is to say, in modelling. Weather
forecasting provides a very telling example of this, one which
clarifies that the main limit to
predictions based on analogs is \emph{not} the sensitivity to initial
conditions, typical of chaos. But, as first realized by Lorenz \cite{CCFV12}
the key difficulty lies with the problem of finding good analogues. 

The first modern steps in weather forecasting are due to 
Richardson \cite{L06} who, in his visionary work, 
introduced many of the ideas on
which modern meteorology is based. His approach was, to a certain
extent, in line with genuine reductionism, and may be summarised as
follows: the atmosphere evolves according to the hydrodynamic
(and thermodynamics) equations for the velocity, the density, and so on.
Therefore, future weather can be predicted, in principle at
least, by solving the proper partial differential equations, 
with initial conditions given by the present state of the atmosphere. 

The key idea by Richardson to forecast the weather was
correct, but in order to put it in practice it was necessary to
introduce one further ingredient that he could not possibly
have known \cite{L06}.
After few decades von Neumann and Charney  noticed that the equations
originally proposed by Richardson, even though correct,
are not suitable for weather forecasting \cite{L06,HV18}.
The apparently
paradoxical reason is that they are too accurate: they also
describe high-frequency wave motions that are irrelevant
for meteorology. So it is necessary to construct effective
equations that get rid of the fast variables. 

The effective equations have great  practical advantages,
e.g. it is possible to adopt large integration time steps  making the numerical computations satisfactorily efficient.
Even more importantly, they are able to capture the essence of the phenomena
of interest, which could otherwise be hidden in too detailed a
description, as in the case of the complete set of original
equations.
It is important to stress  that the effective equations
are not mere approximations of the original
 equations, and  they are obtained with a
subtle mixture of hypotheses, theory and observations \cite{CV17}.
 Just like the cognitive limitations of any human computer provided
Turing with constraints which shaped crucially his model of
computability.

\section{Final remarks and conclusion}

Let us add to the examples discussed in  Sections 3 and 4 two further remarks, which
will lend themselves to a more general conclusion. 

The relevance of the scale resolution is linked with the effective variables describing a phenomenon. 
One understands such a topic considering a  fluid
which can be described in terms of its molecules; in such an approach the
correct variables are the positions and momenta of the molecules. So we
have a very accurate description containing a lot of informations. 
On the other hand, often this  approach  is not interesting, for instance
in engineering (or geophysical) problems we describe a 
fluid in terms of few fields (for velocity, temperature and so on). 
Therefore 
one has a huge decreasing of the amount of information and an increasing of
the possibility to compress data.
\\
Second, a remark on the qualitative aspects of science, which, sometimes,
are considered to be less important than the quantitative
ones. Some results cannot be expressed in terms of numerical sequences, but they
can be interesting and rigorous. 
We can mention the Lotka-Volterra like equations:
it is not possible to  find the explicit solution, but one can show that
the time behavior is periodic. In a similar way, in some celestial mechanics
problems, it is enough to be sure that the motion (e.g. of an asteroid) remain
in a bounded region \cite{CCV10}.
The previous qualitative results, although cannot be formalized in terms
of algorithmic compression (which involves sequences) are genuine forms of
compression of information.

So, here is, in conclusion, the gist of our argument. The claim that the world is comprehensible because it is
algorithmically compressible is, in our opinion, a truism, which is
equivalent to saying that laws of nature exist.  Like many truisms, it
may well have heuristic value. For instance in the familiar situations 
which present a series of numbers which we are invited to continue
appropriately. Guessing rightly the intended continuation of the series, by
all means, is to undestand how the series is generated, i.e. the law
which governs it. But, as we argued above, there are serious
limitations in generalising this approach to the wider and
multifarious field of scientific knowledge. The fact that we know the
law ruling a certain phenomena  does not imply that we are able to
control (e.g. predict) the system. This has been discussed for a
simple example of chaotic system (Bernoulli shift) and for the Navier-Stokes equation for fluids.

Disregarding the distinction between initial conditions and
laws of nature can lead to great confusion. 
For instance  \cite{D90}  claims that chaotic systems
are not algorithmically compressible. 
On the other hand, as discussed in Sect. 2, 
the evolution law ruling chaotic systems
 can be trivially compressible (in the sense that it is easy
to write down the evolution laws, as, e.g., for the Bernoulli's shift). What
can be not compressible is the output and this is related to the complexity
of the sequence associated to the initial condition.

We have pointed out the importance of the  concept of {\it state 
of the system}, i.e. in mathematical terms, the variables which
describe the phenomenon  under investigation, which is often overlooked and
its relevance underestimated. Only in few cases, e.g. in mechanical
systems, it is easy to identify the variables which describe the phenomenon.
On the contrary, in a generic case,  there are serious difficulties;
we can say  that the main effort in building a theory on nontrivial
phenomena  concerns  the identification of the appropriate
variables. 

This difficulty is well-known 
 in the context of  statistical physics:  Onsager and
Machlup, in their seminal work on fluctuations and irreversible processes,
stressed the problem  with the caveat \cite{OM53}.

\begin{quote} {\it how do you know you have taken enough variables,
    for it to be Markovian? }
\end{quote}

In a similar way,  \cite{M85}  notes that

\begin{quote} {\it the hidden worry of thermodynamics is: we do not
    know how many coordinates or forces are necessary to completely
    specify an equilibrium state.}
\end{quote}
Unfortunately, usually we have no definitive method for selecting  the proper
variables and only a deep theoretical understanding can suggest the ``good ones''.
We have recalled that choosing the right question to ask and cutting down the number of variables according to the relevant
practical limitations, were two of the key steps which led Alan
Turing to his model of effective computation. Nearly a century on, the very same ideas
stand still as guidelines for the subtle art of (algorithmic) modelling.

On the other hand, if the laws  are not known and we have just the possibility
to study time series, the scenario is quite pessimistic.
If the effective dimensionality is (relatively) large, 
even in the most simple case of deterministic system, it is not possible to find
the evolution laws and therefore to perform an explicit  compression \cite{CV17}.

The above difficulties have been often underestimated by
the supporters of a science based only on data and algorithms, for a recent
detailed critics of such an  approach which tries to avoid the use of
any theory,   see  \cite{M18,B19}.

\end{document}